# A Conditional Random Field for Discriminatively-trained Finite-state String Edit Distance

**Andrew McCallum and Kedar Bellare**
Department of Computer Science
University of Massachusetts Amherst
Amherst, MA 01003, USA
{mccallum,kedarb}@cs.umass.edu

**Fernando Pereira**
Department of Computer and Information Science
University of Pennsylvania
Philadelphia, PA 10104, USA
pereira@cis.upenn.edu

## Abstract

The need to measure sequence similarity arises in information extraction, object identity, data mining, biological sequence analysis, and other domains. This paper presents *discriminative string-edit CRFs*, a finite-state conditional random field model for edit sequences between strings. Conditional random fields have advantages over generative approaches to this problem, such as pair HMMs or the work of Ristad and Yianilos, because as conditionally-trained methods, they enable the use of complex, arbitrary actions and features of the input strings. As in generative models, the training data does not have to specify the edit sequences between the given string pairs. Unlike generative models, however, our model is trained on both positive and negative instances of string pairs. We present positive experimental results on several data sets.

## 1 Introduction

Parameterized string similarity models based on string edits have a long history (Levenshtein, 1966; Needleman & Wunsch, 1970; Sankoff & Kruskal, 1999). However, there are few methods for *learning* model parameters from training data, even though, as in other tasks, learning may lead to greater accuracy on real-world problems.

Ristad and Yianilos (1998) proposed an expectation-maximization-based method for learning string edit distance with a generative finite-state model. In their approach, training data consists of pairs of strings that should be considered similar, and the parameters are probabilities of certain edit operations. In the E-step, the highest probability edit sequence is found using the current parameters; in the M-step the probabilities are re-estimated using the expectations determined in the E-step so as to reduce the cost of the edit sequences expected to have caused the match. A useful attribute of this method is that the edit operations and parameters can be associated with states of a finite state machine (with probabilities of edit operations depending on previous edit operations, as determined by the finite-state structure.) However, as a generative model, this model cannot tractably incorporate arbitrary features of the input strings, and it cannot benefit from negative evidence from pairs of strings that (while partially overlapping) should be considered dissimilar.

Bilenko and Mooney (2003) extend Ristad's model to include affine gaps, and also present a learned string similarity measure based on unordered bags of words, with training performed by an SVM. Cohen and Richman (2002) use a conditional maximum entropy classifier to learn weights on several sequence distance features. A survey of string edit distance measures is provided by Cohen et al. (2003). However, none of these methods combine the expressive power of a Markov model of edit operations with discriminative training.

This paper presents an undirected graphical model for string edit distance, and a conditional-probability parameter estimation method that exploits both matching and non-matching sequence pairs. Based on conditional random fields (CRFs), the approach not only provides powerful capabilities long sought in many application domains, but also demonstrates an interesting example of discriminative learning of a probabilistic model involving structured latent variables.

The training data consists of input string pairs, each associated with a binary label indicating whether the pair should be considered a "match" or a "mismatch." Model parameters are estimated from both positive and negative examples, unlike in previous generative models (Ristad & Yianilos, 1998; Bilenko & Mooney, 2003). As in those models, however, it is not necessary to provide the desired edit-operations or alignments—the alignments that enable the most accurate discrimi-nation will be discovered automatically through an EM procedure. Thus this model is an example of an interesting class of graphical models that are trained conditionally, but have latent variables, and find the latent variable parameters that maximize discriminative performance. Another recent example includes work on CRFs for object recognition from images (Quattoni et al., 2005).

The model is structured as a finite-state machine (FSM) with a single initial state and two disjoint sets of non-initial states with no transitions between them. State transitions are labeled by edit operations. One of the disjoint sets represents the match condition, the other the mismatch condition. Any non-empty transition path starting at the initial state defines an edit sequence that is wholly contained in either the match or mismatch subsets of the machine. By marginalizing out all the edit sequences in a subset, we obtain the probability of match or mismatch.

The cost of a transition is a function of its edit operation, the previous state, the new state, the two input strings, and the starting and ending position (the position of the match-so-far before and after performing this edit operation) for each of the two input strings. In applications, we take full advantage of this flexibility. For example, the cost function can examine portions of the input strings both before and after the current match position, it can examine domain knowledge, such as lexicons, or it can depend on rich conjunctions of more primitive features.

The flexibility of edit operations is possibly even more valuable. Edits can make arbitrarily-sized forward jumps in both input strings, and the size of the jumps can be conditioned on the input strings, the current match points in each, and the previous state of the finite state process. For example, a single edit operation could match a three-letter acronym against its expansion in the other string by consuming three capitalized characters in the first string, and consuming three matching words in the second string. The cost of such an operation could be conditioned on the previous state of the finite state process, as well as the appearance of the consumed strings in various lexicons, and the words following the acronym.

Inference and training in the model depends on a complex dynamic program in three dimensions. We employ various optimizations to speed learning.

We present experimental results on five standard text data sets, including short strings such as names and addresses, as well as longer more complex strings, such as bibliographic citations. We show significant error reductions in all but one of the data sets.

## 2 Discriminatively Trained String Edit Distance

Let $\mathbf{x} = x_1 \cdots x_m$ and $\mathbf{y} = y_1 \cdots y_n$ be two strings or symbol sequences. This pair of input strings is associated with an output label $z \in \{0, 1\}$ indicating whether or not the strings should be considered a match (1) or a mismatch (0).[1] As we now explain, our model scores *alignments* between $\mathbf{x}$ and $\mathbf{y}$ as to whether they are a match or a mismatch. An alignment $\mathbf{a}$ is a four-tuple consisting of a sequence of edit operations, two sequences of string positions, and a sequence of FSM states.

Let $\mathbf{a}.\mathbf{e} = e_1 \cdots e_k$ indicate the sequence *edit operations*, such as delete-one-character-in-x, substitute-one-character-in-x-for-one-character-in-y, or delete-all-characters-in-x-up-to-its-next-nonalphabetic. Each edit operation $e_p$ in the sequence consumes either some of $\mathbf{x}$ (deletion), some of $\mathbf{y}$ (insertion), or some of both (substitution), up to positions $i_p$ in $\mathbf{x}$ and $j_p$ in $\mathbf{y}$. We have therefore corresponding non-decreasing sequences $\mathbf{a}.\mathsf{ix} = i_1, \ldots, i_k$ and $\mathbf{a}.\mathsf{iy} = j_1, \ldots, j_k$ of edit-operation positions for $\mathbf{x}$ and $\mathbf{y}$.

To classify alignments into matches or mismatches, we take edits as transition labels for a non-deterministic FSM with state set $S = \{q_0\} \cup S_0 \cup S_1$. There are transitions from the initial state $q_0$ to states in the disjoint sets $S_0$ and $S_1$, but no transitions between those two sets. In addition to the edit sequence and string position sequences, we associate the alignment $\mathbf{a}$ with a sequence of consecutive destinations states $\mathbf{a}.\mathsf{q} = q_1 \cdots q_k$, where $e_p$ labels an allowed transition from $q_{p-1}$ to $q_p$. By construction, either $\mathbf{a}.\mathsf{q} \subseteq S_0$ or $\mathbf{a}.\mathsf{q} \subseteq S_1$. Alignments with states in $S_1$ are supposed to represent matches, while alignments with states in $S_0$ are supposed to represent mismatches.

In summary, an alignment is specified by the four-tuple $\mathbf{a} = \langle \mathbf{a}.\mathbf{e} = e_1 \cdots e_k, \mathbf{a}.\mathsf{ix} = i_1 \cdots i_k, \mathbf{a}.\mathsf{iy} = j_1 \cdots j_k, \mathbf{a}.\mathsf{q} = q_1 \cdots q_k \rangle$. For convenience, we also write $\mathbf{a} = a_0, a_1 \cdots a_k$ with $a_p = \langle e_p, i_p, j_p, q_p \rangle, 1 \leq p \leq k$ and $a_0 = \langle -, 0, 0, q_0 \rangle$ where $-$ is a dummy initial edit.

Given two strings $\mathbf{x}$ and $\mathbf{y}$, our discriminative string edit CRF defines the probability of an alignment between $\mathbf{x}$ and $\mathbf{y}$ as

$$p(\mathbf{a}|\mathbf{x}, \mathbf{y}) = \frac{1}{Z_{\mathbf{x},\mathbf{y}}} \prod_{i=1}^{|\mathbf{a}|} \Phi(a_{i-1}, a_i, \mathbf{x}, \mathbf{y}),$$

---

[1] One could also straightforwardly imagine a different regression-based scenario in which $z$ is real-valued, or also a ranking-based criteria, in which *two* pairs are provided and $z$ indicates which pair of strings should be considered closer.

where the potential function $\Phi(\cdot)$ is a non-negative function of its arguments, and $Z_{\mathbf{x},\mathbf{y}}$ is the normalizer (partition function). In our experiments we parameterize these potential functions as an exponential of a linear scoring function

$$\Phi(a_{i-1}, a_i, \mathbf{x}, \mathbf{y}) = \exp \Lambda \cdot \mathbf{f}(a_{i-1}, a_i, \mathbf{x}, \mathbf{y}),$$

where $\mathbf{f}$ is a vector of *feature functions*, each taking as arguments two consecutive states in the alignment sequence, the corresponding edits, and their string positions, which allow the feature functions to depend on the context of $a_i$ in $\mathbf{x}$ and $\mathbf{y}$. A typical feature function combines some predicate on the input, or *input feature*, with a predicate over the alignment itself (edit operation, states, positions).

To obtain the probability of match given simply the input strings, we marginalize over all alignments in the corresponding state set:

$$p(z|\mathbf{x}, \mathbf{y}) = \sum_{\mathbf{a}.\mathbf{q} \subseteq S_z} \frac{1}{Z_{\mathbf{x},\mathbf{y}}} \prod_{i=1}^{|\mathbf{a}|} \Phi(a_{i-1}, a_i, \mathbf{x}, \mathbf{y}),$$

Fortunately, this sum can be calculated efficiently by dynamic programming. Typically, for any given edit operation, starting positions and input strings, there are a small number of possible resulting ending positions. Max-product (Viterbi-like) inference can also be performed efficiently.

## 3 Parameter Estimation

Parameters are estimated by penalized maximum likelihood on a set of training data. Training data consists of a set of $N$ string pairs $\langle \mathbf{x}^{(j)}, \mathbf{y}^{(j)} \rangle$ with corresponding labels $z^{(j)} \in \{0, 1\}$, indicating whether or not the pair is a match. We use a zero-mean spherical Gaussian prior $\sum_k \lambda_k^2 / \sigma^2$ for penalization.

The incomplete (non-penalized) log-likelihood is then

$$\mathcal{L}_I = \Big( \sum_j \log p(z^{(j)}|\mathbf{x}^{(j)}, \mathbf{y}^{(j)}) \Big)$$

and the complete log-likelihood is

$$\mathcal{L}_C = \Big( \sum_j \sum_{\mathbf{a}} \log(p(z^{(j)}|\mathbf{a}, \mathbf{x}^{(j)}, \mathbf{y}^{(j)}) p(\mathbf{a}|\mathbf{x}^{(j)}, \mathbf{y}^{(j)})) \Big)$$

We maximize this likelihood with EM, estimating $p(\mathbf{a}|\mathbf{x}^{(j)}, \mathbf{y}^{(j)})$ given current parameters $\Lambda$ in the E-step, and maximizing the complete penalized log-likelihood in the M-step. For optimization in the M-step we use BFGS. Unlike CRFs without latent variables, the objective function has local maxima. To avoid getting stuck in poor local maxima, the parameters are initialized to yield a reasonable default edit distance.

Dynamic programming for this model fills a three-dimensional table (two for the two input strings, and one for the states in $S$). The table can be moderately large in practice ($n = m = 100$ and $|S| = 12$, resulting in 120,000 entries), and beam search may effectively be used to increase speed, just as in speech recognition, where even larger tables are common.

It is interesting to examine what alignments will be learned in $S_0$, the non-match portion of the model. To attain high accuracy, these states should attract string pairs that are dissimilar. But even similar strings have bad alignments, for example the alignment that first deletes all of $\mathbf{x}$, and then inserts all of $\mathbf{y}$. Fortunately, finding *how dissimilar* two strings are requires finding as good an alignment as is possible, and then deciding that this alignment is not very good. These as-good-as-possible alignments are exactly what our learning procedure discovers: driven by an objective function that aims to maximize the likelihood of the correct binary match/non-match labels, the model finds the latent alignment paths that enable it to maximize this likelihood.

This model thus falls in a family of interesting techniques involving discrimination among complex structured objects, in which the structure or relationship among the parts is unknown (latent), and the latent choice has high impact on the discrimination task. Similar considerations are at the core of discriminative non-probabilistic methods for structured problems such as handwriting recognition (LeCun et al., 1998) and speech recognition (Woodland & Povey, 2002), and, more recently, computer vision object recognition (Quattoni et al., 2005). We discuss related work further in Section 6.

## 4 Implementation

The model has been implemented as part of the finite-state transducer classes in MALLET (McCallum, 2002). We map three-dimensional dynamic programming problems over positions in $\mathbf{x}$ and $\mathbf{y}$ and states $S$ to MALLET's existing finite-state forward-backward and Viterbi implementations by encoding the two position indices into a single index in a diagonal crossing pattern that starts at $(0, 0)$. For example, a single-character delete operation, which would be a hop to an a adjacent vertical or horizontal in the original table, is a longer, one-dimensional (but deterministically-calculated) jump in the encoding.

In addition to the standard edit operations (insertion, deletion, substitution), we have also more powerful edits that fit naturally into this model, such as delete-until-end-of-word, delete-word-in-lexicon, and delete-word-appearing-in-other-string.

# 5 Experimental Results

We show experimental results on one synthetic and six real-world data sets, all of which have been used in previous work evaluating string edit measures. The first two data sets are the name and address fields of the *Restaurant* database. Among its 864 records, 112 are matches. The last four data sets are citation strings from the standard *Reasoning, Constraint, Reinforcement* and *Face* sections of the *CiteSeer* data. The ratios of citations to unique papers for these are 514/196, 349/242, 406/148 and 295/199 respectively. Making the problem more challenging than certain other evaluations on these data sets, our strings are *not* segmented into fields such as title or author, but are each treated as a single unsegmented character sequence. We also present results on synthetic noise on person names, generated by the UIS Database generator. This program produces perturbed names according to modifiable noise parameters, including the probability of an error anywhere in a record, the probability of single character insertion, deletion or swap, and the probability of a word swap.

## 5.1 Edit Operations and Features

One of the main advantages of our model is the ability to include non-independent input features and extremely flexible edit operations. The input features used in our experiments include subsets of the following, described as acting on cell $i, j$ in the dynamic programming table and the two input strings $\mathbf{x}$ and $\mathbf{y}$.

- same, different : $x_i$ and $y_j$ match (do not match);
- same-alphabetic, different-alphabetic : $x_i$ and $y_j$ are alphabetic and they match (do not match);
- same-numeric, different-numeric : $x_i$ and $y_j$ are numeric and they match (do not match);
- punctuation-x, punctuation-y : $x_i$ and $y_j$ are punctuation, respectively;
- alphabet-mismatch, number-mismatch : One of $x_i$ and $y_j$ is alphabetic (numeric), the other is not;
- end-of-x, end-of-y : $i = |\mathbf{x}|$ $(j = |\mathbf{y}|)$;
- same-next-character, different-next-character: $x_{i+1}$ and $y_{i+1}$ match (do not match).

Edit operations on FSM transitions include:

- Standard string edit operations: insert, delete and substitute.
- Two character operations: swap-two-characters.
- Word skip operations: skip-if-word-in-lexicon, skip-word-if-present-in-other-string, skip-parenthesized-words and skip-any-word .
- Operations for handling acronyms and abbreviations by inserting, deleting, or substituting specific types of substrings.

Learned parameters are associated with the input features as well as with state transitions in the FSM. All transitions entering a state may share tied parameters (first order), or have different parameters (second order). Since the FSM can have more states than edit operations, it can remember the context of previous edit actions.

## 5.2 Experimental Methodology

Our model exploits both positive and negative examples during training. Positive training examples include all pairs of strings referring to the same object (the matching strings). However, the total number of negative examples is quadratic in the number of objects. Due to both time and memory constraints, as well as a desire to avoid overwhelming the positive training examples, we sample the negative (mismatch) string pairs so as to attain a 1:10 ratio of match to mismatch pairs. In order to preferentially sample "near misses" we filter negative examples in one of two ways:

- Remove negative examples that are too dissimilar according to a suitable metric. For the *Citeseer* datasets we use the cosine metric to measure similarity of two citations; for other datasets we use the metric of Jaro (1989).

- Select the best matching negative pairs according to a CRF with parameters set by hand to reasonable values.

As in Bilenko and Mooney (2003), we use a 50/50 train/test split of the data, and repeat the process with the folds interchanged. With the restaurant name and restaurant address dataset, we run our algorithm with different choices of features and states, and 4 random splits of the data. With the *Citeseer* datasets, we have results for two random splits of the data.

To give EM training a reasonable starting point, we hand-set the initial parameters to somewhat arbitrary, yet reasonable parameters. (Of course, hand-setting of string edit parameters is the standard for all the non-learning approaches.) We examined a small held-out set of data to verify that these initial parameters were reasonable. We set the parameters on the match portion of the FSM to provide good alignments; then we then copy these parameters to the mismatch portion of the model, offsetting them by bringing all values closer to zero by a small constant.

| Distance Metric | Restaurant name | Restaurant address | Reasoning | Face | Reinforcement | Constraint |
|---|---|---|---|---|---|---|
| Edit Distance | 0.290 | 0.686 | 0.927 | 0.952 | 0.893 | 0.924 |
| Learned Edit Distance | 0.354 | 0.712 | 0.938 | **0.966** | 0.907 | 0.941 |
| Vector-space | 0.365 | 0.380 | 0.897 | 0.922 | 0.903 | 0.923 |
| Learned Vector-space | 0.433 | 0.532 | 0.924 | 0.875 | 0.808 | 0.913 |
| CRF Edit Distance | **0.448** | **0.783** | **0.964** | 0.918 | **0.917** | **0.976** |

Table 1: Averaged F-measure for detecting matching field values on several standard data sets (bold indicates highest F1). The top four rows are results duplicated from Bilenko and Mooney (2003); the bottom row is the performance of the CRF method introduced in this paper.

Lexicons were populated automatically by gathering the most frequent words in the training set. (Alternatively one could imagine lexicon feature values set to inverse-document-frequency values, or similar information retrieval metrics.) In some cases, before training, lexicons were edited to remove author surnames.

The equations in section 3 are used to calculate $p(z|\mathbf{x}, \mathbf{y})$, with a first-order model. A threshold of 0.5 predicts whether the string pair is a match or a mismatch. (Note that alternative thresholds could easily be used to trade of precision and recall, and that CRFs are typically good at predicting calibrated posterior probabilities needed for such tuning as well as accuracy/coverage curves.) Bilenko and Mooney (2003) found transitive closure to improve F1, and use it for their results; we did not find it to help, and do not.

Precision is calculated to be the ratio of the number of correctly classified duplicates to the total number of duplicates identified. Recall is the ratio of correctly classified duplicates to the total number of duplicates in the dataset. We report the mean performance across multiple random splits.

### 5.3 Results

In experiments on the six real-world data sets we compare our performance against results in a recent benchmark paper by Bilenko and Mooney (2003); Bilenko recently completely thesis work in this area. These results are summarized in Table 1, where the top four rows are duplicated from Bilenko and Mooney (2003), and the bottom row shows the results of our method. The entries are the average F1 measure across the folds. We observe large performance improvements on most datasets. The fact that the difference in performance across our trials is typically around 0.01 suggests strong statistical significance. Our average F1 on the Face dataset was 0.04 less than the previous best. The examples on which we made errors generally had a large venue, authors, or URL field in one string but not in the other.

We also evaluate the effect on performance of using Viterbi (max-product) inference in training instead of forward-backward (sum-product) inference. Except for the restaurant address dataset, forward-backward performs significantly better than Viterbi on all datasets. The restaurant address data set contains positive examples with a large unmatched suffix in one of the strings, which may lead to an inappropriate dilution of probability amongst many alignments. Average F1 measures for the restaurant datasets using Viterbi and forward-backward are shown in Table 2. All results shown in Table 1 use forward-backward probabilities.

| Dataset | Viterbi | Forward-Backward |
|---|---|---|
| Restaurant name | 0.689 | 0.720 |
| Restaurant address | 0.708 | 0.651 |

Table 2: Averaged F-measures for Viterbi vs. forward-backward on (trained and evaluated on a subset of the data; smaller test set yields higher accuracy).

In the other tables we present results showing the impact of various edit operations and features.

Table 3 shows F1 on the restaurant data set as various edit operations are added to the model: $i$ denotes insert, $d$ denotes delete, $s$ denotes substitute, paren denotes skip-parenthesized-word, lex denotes skip-if-word-in-lexicon, and pres denotes skip-word-if-present-in-other-string. All use the same-alphabets and different-alphabets input features. As can be seen from the results, adding "skip" edits improves performance. Although skip-parenthesized-words gives better results on the smaller data set used for the experiments in the table, skip-if-word-in-lexicon produces a higher accuracy on larger data sets, because of peculiarities in how restaurants with the same name and different locations are named in the data set. We also see that a second-order model performs less well, presumably because of data sparseness.

Table 4 shows the benefits of including various features for the restaurant address data set, while fixing the edit operations (insert, delete and substitute). In the table, $s$ and $d$ denote the same and different features, salp and *dalp* stand for the same-alphabets and different-alphabets features, and *snum* and *dnum* stand for the same-numbers and different-numbers features. The *s* and *d* features are different from the *salp*, *dalp*, *snum*, and *dnum* features in that the weights learned for the former depend only on whether the two characters are equal or not, and no separate weights are learned for a number match or an letter match. We conjecture that a number mismatch in the address data needs to be penalized more than a letter mismatch. Separating the same and different features into features for letters and numbers reduces the error from about 6% to 3%.

| Run | F1 |
|---|---|
| $i, d, s$ | 0.701 |
| $i, d, s, paren$ | 0.835 |
| $i, d, s, lex$ | 0.769 |
| $i, d, s, lex\ 2^{nd}$ order | 0.742 |
| $i, d, s, paren, lex, pres$ | 0.746 |
| $i, d, s, paren, lex, pres, 2^{nd}$ order | 0.699 |

Table 3: Averaged maximum F-measure for different state combinations on a subset of restaurant name (trained and evaluated on the same train/test split).

| Run | F1 |
|---|---|
| $s, d$ | 0.944 |
| $salp, dalp, snum, dnum$ | 0.973 |

Table 4: Averaged maximum F1-measure for different feature combinations on a subset of the restaurant address data set.

Finally, Table 5 demonstrates the power of CRFs to include extremely flexible edit operations that examine arbitrary pieces of the two input strings. In particular we measure the impact of including the skip-word-if-present-in-other-string operation, ("skip" for short). Here we train and test on the UIS synthetic name data, in which the error probability is 40%, the typo error probability is 40% and the swap first and last name probability is 50%; (the rest of the parameters were unchanged from the default values). The difference in performance is dramatic, bringing error down from about 14% to less than 2%. Of course, arbitrary substring swaps are not expressible in standard dynamic programs, but the skip operation gives an excellent approximation while preserving efficient finite-state inference. Typical improved alignments with the new operation may skip over a matching swapped first name, and then proceed to correct individual typographic errors in the last name.

An example alignment found by our model on restaurant name is shown in Table 7. As discussed in Sec-tion 3, the mismatch portion of the model indeed learns the best possible latent alignments in order to measure distance with the most salient features. This example's alignment score from the match portion is higher. The entries in the dynamic programming table i, d, s, l, and p correspond to states reached by the operations insert, delete, substitute, skip-word-in-lexicon, and skip-parenthesized-word respectively. The symbol - denotes a null transition.

| Run | F1 |
|---|---|
| Without *skip* | 0.856 |
| With *skip* | 0.981 |

Table 5: Average maximum F-measure for synthetic name dataset with and without skip-if-present-in-other-string state.

|   | ε | r | e | s | t | a | u | r | a | n | t | : | k | a | t | z | u | ε |
|---|---|---|---|---|---|---|---|---|---|---|---|---|---|---|---|---|---|---|
| ε | l |   |   |   |   |   |   |   |   |   |   |   |   |   |   |   |   |   |
| k |   |   |   |   |   |   |   |   |   |   |   | s |   |   |   |   |   |   |
| a |   |   |   |   |   |   |   |   |   |   |   |   | s |   |   |   |   |   |
| t |   |   |   |   |   |   |   |   |   |   |   |   |   | s |   |   |   |   |
| s |   |   |   |   |   |   |   |   |   |   |   |   |   |   | **s** |   |   |   |
| u |   |   |   |   |   |   |   |   |   |   |   |   |   |   |   | s |   |   |
| ε |   |   |   |   |   |   |   |   |   |   |   |   |   |   |   |   | s | - |

Table 6: Alignment in both the match and mismatch subsets of the model, with correct prediction. Operations causing edits are in bold.

Table 7: Alignment in both the match and mismatch subsets of the model, with correct prediction. Operations causing edits in bold.

## 6 Related Work

String (dis)similarity metrics based on edit distance are widely used in applications ranging from approximate matching and duplicate removal in database records to identifying conserved regions in comparative genomics. Levenshtein (1966) introduced least-cost editing based on independent symbol insertion, deletion, and substitution costs, and Needleman and Wunsch (1970) extended the method to allow gaps. Editing between strings over the same alphabet can be generalized to transduction between strings in different alphabets, for instance in letter-to-sound mappings (Riley & Ljolje, 1996) and in speech recognition (Jelinek et al., 1975).

In most applications, the edit distance model is derived by heuristic means, possibly including some data-dependent tuning of parameters. For example, Monge and Elkan (1997) recognize duplicate corrupted records using an edit distance with tunable edit and gap costs. Hernandez and Stolfo (May 1995) merge records in large databases using rules based on domain-specific edit distances for duplicate detection. Cohen (2000) use a token-based TF-IDF string similarity score to compute ranked approximate joins on tables derived from Web pages. Koh et al. (2004) use association rule mining to check for duplicate records with per-field exact, Levenshtein or BLAST 2 gapped alignment (Altschul et al., 1997) matching. Cohen et al. (2003) surveys edit and common substring similarity metrics for name and record matching, and their application in various duplicate detection tasks.

In bioinformatics, sequence alignment with edit costs based on evolutionary or biochemical estimates are common (Durbin et al., 1998). Position-independent costs are normally used for general sequence similarity search, but position-dependent costs are often used when searching for specific sequence motifs.

In basic edit distance, the cost of individual edit operations is independent of the string context. However, applications often require edit costs to change depending on context. For instance, the characters in an author's first name after the first character are more likely to be deleted than the first character. Instead of specialized representations and dynamic programming algorithms, we can instead represent context-dependent editing with weighted finite-state transducers (Eilenberg, 1974; Mohri et al., 2000) whose states represent different types of editing contexts. The same idea has also been expressed with pair hidden Markov models for pairwise biological sequence alignment (Durbin et al., 1998).

If edit costs are identified with $-\log$ probabilities (up to normalization), edit distance models and certain weighted transducers can be interpreted as generative models for pairs of sequences. Pair HMMs are such generative models by definition. Therefore, expectation-maximization using an appropriate version of the forward-backward algorithm can be used to learn parameters that maximize the likelihood of a given training set of pairs of strings according to the generative model (Ristad & Yianilos, 1998; Ristad & Yianilos, 1996; Durbin et al., 1998). Bilenko and Mooney (2003) use EM to train the probabilities in a simple edit transducer for one of the duplicate detection measures they evaluate. Eisner (2002) gives a general algorithm for learning weights for transducers, and notes that the approach applies to transducers with transition scores given by globally normalized log-linear models. These models are to CRFs as pair HMMs are to HMMs.

The foregoing methods for training edit transducers or pair HMMs use positive examples alone, but do not need to be given explicit alignments because they do EM with alignment as a latent (structured) variable. Joachims (2003) gives a generic maximum-margin method for learning to score alignments from positive and negative examples, but the training examples must include the actual alignments. In addition, he cannot solve the problem exactly because he does not exploit factorizations of the problem that yield a polynomial number of constraints and efficient dynamic programming search over alignments.

While the basic models and algorithms are expressed in terms of single letter edits, in practice it is convenient to use a richer application-specific set of edit operations, for example name abbreviation. For example, Brill and Moore (2000) use edit operations designed for spelling correction in a spelling correction model trained by EM. Tejada et al. (2001) has edit operations such as abbreviation and acronym for record linkage.

# 7 Conclusions

We have presented a new discriminative model for learning finite-state edit distance from postive and negative examples consisting of matching and non-matching strings. It is not necessary to provide sequence alignments during training. Experimental results show the method to outperform previous approaches.

The model is an interesting member of a family of models that use a discriminative objective function to discover latent structure. The latent edit operation sequences that are learning by EM are indeed the alignments that help discriminate matching from non-matching strings.

We have described in some detail the finite-state version of this model. A context-free grammar version of the model could, through edit operations defined on trees, handle swaps of arbitrarily-sized substrings.

# Acknowledgments

We thank Charles Sutton and Xuerui Wang for useful conversations, and Mikhail Bilenko for helpful comments on a previous draft. This work was supported in part by the Center for Intelligent Information Retrieval, the National Science Foundation under NSF grants #IIS-0326249, #IIS-0427594, and #IIS-0428193, and by the Defense Advanced Research Projects Agency, through the Department of the Interior, NBC, Acquisition Services Division, under contract #NBCHD030010.